# Saliency Fusion in Eigenvector Space with Multi-Channel Pulse Coupled Neural Network

Nevrez İmamoğlu, Zhixuan Wei, Huangjun Shi, Yuki Yoshida, Myagmarbayar Nergui, Jose Gonzalez, Dongyun Gu, Weidong Chen, Kenzo Nonami, Wenwei Yu

*Abstract*—**Saliency computation has become a popular research field for many applications due to the useful information provided by saliency maps. For a saliency map, local relations around the salient regions in multi-channel perspective should be taken into consideration by aiming uniformity on the region of interest as an internal approach. And, irrelevant salient regions have to be avoided as much as possible. Most of the works achieve these criteria with external processing modules; however, these can be accomplished during the conspicuity map fusion process. Therefore, in this paper, a new model is proposed for saliency/conspicuity map fusion with two concepts: a) input image transformation relying on the principal component analysis (PCA), and b) saliency conspicuity map fusion with multi-channel pulsed coupled neural network (*m*-PCNN). Experimental results, which are evaluated by *precision*, *recall*, *F-measure*, and area under curve (AUC), support the reliability of the proposed method by enhancing the saliency computation.**

*Index Terms*—**information fusion, multi-channel pulse coupled neural network, principal component analysis, saliency detection.**

## I. INTRODUCTION

INFORMATION fusion is a wide and important topic for industrial applications and engineering field such as robotics, biomedical applications, computer vision, etc. Fusion can be seen in two ways based on the input variations; a) integration of different sensors in which the sensor data types may differ [1] b) fusion of the same type of data or features generated from the same type of source such as images or image features [2-5]. In addition, as stated by Luo and Chang [1], the grouping can be made based on the fusion algorithm as a different perspective: a) low level (estimation methods), b) medium level (classification methods), and c) high level (inference methods) fusion [1].

In this paper, our concern is similar to the fusion of image feature maps created from the same sensor type data or source (e.g. the MRI images [2], RGB color images [4], SAR images [6], active mm-wave images [7]), and the fusion algorithm is in medium level according to the method base grouping. Therefore, the main goal is to create a fused data with more informative data that could be used for further processing such as detection, classification, segmentation, noise reduction, data reduction, etc.

In this study, saliency conspicuity maps are employed for fusion to create a final saliency map. Saliency computation is one of the recent research areas evolved from the concept of visual attention, especially on the human visual system [8, 9]. The visual attention structure provides features by selecting significant parts and reducing the redundant data from the perceived visual scene [10]. This property of the visual attention has been attracting researchers' attention to create computational models for computer vision applications [11, 12]. The general idea is to make the significant features more visible by combining several saliency conspicuity maps (intensity, color, orientation, motion, etc.) into a final saliency map which depicts the attentive regions locally and globally on the image.

For a saliency computation model, local uniformity and continuity around salient points are crucial to detect the salient object region correctly. Also, the fusion method for the feature maps or conspicuity maps should consider not only the pixel to pixel integration of the same locations but also the relation around the surrounding pixels regarding the saliency values. And, local and global saliency relations for the given feature space are also important for any saliency application. In addition, fusion of the feature maps or conspicuity maps is generally based on summation, maximum value selection, or statistical approaches where the connectionist relation between a pixel and its surrounding neighbors through the several salient conspicuity maps can be missing. This will cause irrelevant salient regions or non-salient regions on the object of interest to occur on the saliency map. In many applications, these problems are tried to be solved by internal or external approaches to the saliency computation such as the normalization of feature maps (internal approach) [11] and segmentation (external approach) to enhance salient object detection [12]. The properties of the external approaches can be referred as: they depend on application, need to be fine tuned towards application, and are affected by the implementation details of application. Considering the internal approaches, they are independent of the application, can utilize space projection, feature extraction, fusion concepts as a closed approach in an integrated manner, are easy to be used as a module and can be embedded in to a big application. Therefore, in this paper, we propose an internal approach that refers to the use of principal component analysis (PCA) [13] and multi-channel pulse coupled neural network (*m*-PCNN) fusion [2] with one of the most recent saliency computation method based on Wavelet transform (WT) [14] to enhance the saliency map during fusion process. The local saliency model in [14] has saliency features from edge to textures as a good representation of saliency with local to global contrast for each feature channel. Therefore, it is a good candidate for integrating the proposed model.

PCA is included in many applications regarding saliency map computation [15, 16] or information/data fusion [6, 7]. For saliency applications, the common way is to utilize PCA on image patches to create a feature space with or without reducing the dimension by selecting the features, and then these features can be used to create saliency maps according to the dissimilarities among the image patches [15] or decision on clusters [16]. On the other hand, PCA can also be applied on the image by selecting some specific principal component transforms (PCTs) to be fused by averaging as in [7]. In this study, PCTs on an RGB image is employed to obtain three different 2D images in which the RGB image will be represented by $1^{st}$ to $3^{rd}$ PCTs where local to global information distribution can also be observed through PCTs. And, local saliency computation [14] is applied to each transformed channel separately to generate three saliency conspicuity maps to be fused for the saliency instead of the patch based PCA method. These enable us to control the local and global saliency content by using the eigenvalues of each principle components during $m$-PCNN fusion of saliency conspicuity maps obtained from the PCTs and WT based saliency.

PCNN is an unsupervised neural network structure which has various usages in image processing areas like segmentation, feature extraction, recognition, and fusion [17]. As stated before, the fusion models are generally for different type of images rather than saliency conspicuity map integration [17, 18]. Most of the PCNN applications related to saliency computation are generally for segmentation and enhancement processes after the saliency output is already obtained [19], or to create attention shifts while producing saliency from one channel [20]. On the other hand, most of the multi-channel models are based on many PCNN modules working separately to fuse their outputs [18]. This inefficient method is improved by Wang and Ma's work [2] which proposes $m$-PCNN method to fuse multi-channel input in one PCNN module. However, there is not any saliency conspicuity map fusion based on $m$-PCNN in eigenvector space. Therefore, in this study, it is shown that $m$-PCNN can also be utilized during conspicuity map fusion enhancing the saliency map for multi-channel cases. Moreover, we have enhanced the model by integrating automatic weighting of the input channels with their respective eigenvalues obtained from PCA. Hence, local and global content of the fused saliency map can be weighted for each test image differently. The experimental results demonstrate that the saliency computation model [14] based on the centre-surround difference in multi-resolution analysis is improved with the integration of PCA and $m$-PCNN.

The paper is organized as follows: Section II describes the proposed model, Section III includes the experimental results and comparisons with discussion, and finally conclusion statements are given in Section 4.

## II. PROCEDURE FOR THE MODEL

The proposed method inputs three transformed image channels from an RGB image by using the three eigenvectors obtained by PCA. Then, saliency feature maps are generated with WT based local saliency generator to demonstrate the effectiveness of the PCA and $m$-PCNN integration to an existing state of the art saliency map algorithm. As a final step, $m$-PCNN is utilized for the fusion of three generated saliency conspicuity maps that are weighted with their respective eigenvalues. The procedure of the proposed system can be shown as in Fig.1:

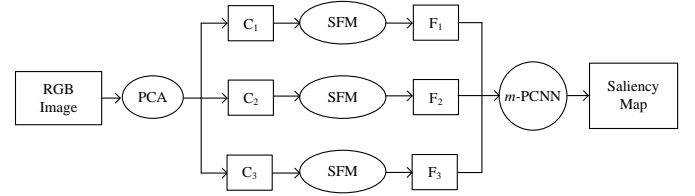

Fig. 1. Flowchart of the proposed model

where $C_i$ is the transformed data on the eigenvector space for each principal components, SFM is the saliency conspicuity map generation, $F_i$ is the saliency conspicuity map, $m$-PCNN is the fusion model based on the multi-channel pulse coupled neural network.

### A. Input Image Transformation by PCA

The initial part of the proposed system is to have the PCA for an *RGB* input image as in [13] to obtain the respective eigenvalues and eigenvectors. Firstly, a Gaussian filter is applied to remove noisy data from the image, and then, the *RGB* image is converted to two dimensional data by making each *R*, *G*, and *B* channel a raw vector. Each channel is adjusted to have zero mean. Now, we have a 2D data **v** where the row size is the size of the image and column size is three, in which each column represents the $R$, $G$ and $B$ color channels. The eigenvalues and eigenvectors are obtained by the covariance matrix of $\mathbf{v}$. And, the 1st principal component as for eigenvalues and eigenvector pairs can be selected as the highest eigenvalues among the set. The 2nd and 3rd principal components are also ordered with respect to their relative eigenvalues.

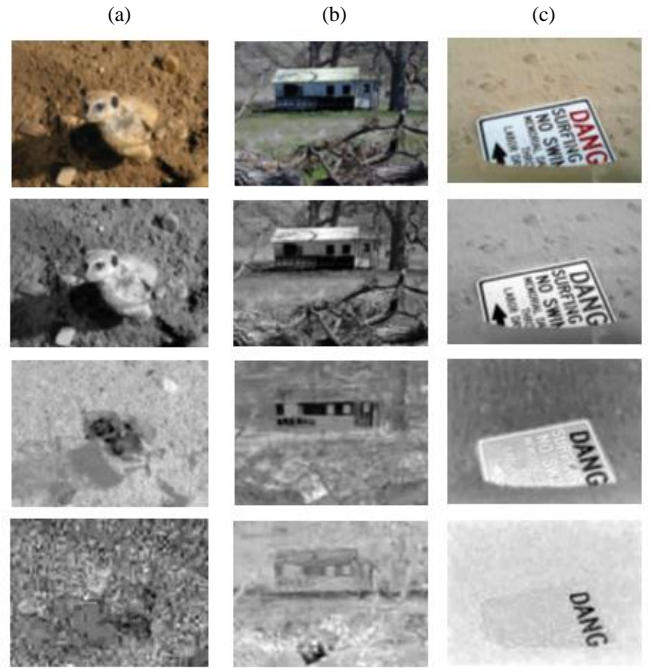

Fig. 2. Sample RGB images and their respective eigenvector space representations ($1^{st}$, $2^{nd}$, and $3^{rd}$ principal transformations respectively)

The transformations for each eigenvector space can be done as in (1) [13]:

$$A_i = (\xi_i^T \times \mathbf{v}^T);\qquad(1)$$

where $A_i$ is the transformed data on the eigenvector space based on the $i^{th}$ principal component, $\mathbf{v}$ is the zero mean adjusted data, $\xi_i$ is the $i^{th}$ principal eigenvector, and $T$ is the transpose operation. And then, 1D $A_i$ is remapped to 2D $C_i$ data regarding the row and column size of the image where $C_i$ is the input images for the saliency computation.

Eigenvector space of the 1st principal component will have the highest variation, and it is the most representative distribution of the data [13]. On the other hand, the transformed data of the 3rd principal component has the least variation [13]. Thus, we can state that the transformed data on the 1st eigenvector space has more local information, and local information will lessen through the 3rd eigenvector space. In Fig.2, some *RGB* color images are given with their 1st, 2nd, and 3rd principal eigenvector space representations from top to bottom. As it can be seen, a more global saliency result can be computed from the 3rd transformation even if local centre-surround model is applied. In Fig.2 (c), the example shows that the red letters (written as "*DANG*") are globally more salient and can be extracted from the 3rd principal component transformed data. On the other hand, the 1st principal component transformation has more local salient information compared to 2nd and 3rd transformed input channels. Obviously, local and global saliency information content can be controlled with PCA based representation instead of using other tools such as feature normalization methods [11] or external assistive segmentation algorithm [12] to enhance the saliency map.

### B. Saliency Conspicuity Maps for each Component

For more detailed information, it is necessary to have saliency conspicuity maps with full resolution where center-surround differences can be observed for each location on the input image. Since Wavelet decomposition's detail signals provide these regional differences from edge to texture, it is a useful tool for saliency computation [14]. In addition, it is good to have local to global saliency model with more local importance (i.e. multi-resolution centre-surround differences) to show the performance of PCA and *m*-PCNN where the *m*-PCNN aims to improve final saliency result by considering the relation among the neighbor pixels during the fusion process. Therefore, the local saliency model of the previous work [14] is adopted for the proposed model to compute the saliency conspicuity maps of each input channel separately as in Fig.1.

The saliency computation of each input channel starts with the wavelet decomposition with several levels depending on the size of the image (2) [14]. The decomposition yields an approximate signal and detail signals with the horizontal, vertical and diagonal cases in several sub-bands [14]. The wavelet decomposition for various levels can be expressed as follows [14]:

$$[\mathbf{A}_N^c, \mathbf{H}_s^c, \mathbf{V}_s^c, \mathbf{D}_s^c] = WT_N\left(\mathbf{I}^c\right)\qquad(2)$$

where $\mathbf{I}^c$ is the input data channel for each 1st, 2nd, and 3rd principal component transformation and normalized to {0-255}, $WT(.)$ is the wavelet decomposition function for the input channel $c$ and scaling level $s$ such that $s \in \{1,\ldots,N\}$, $N$ is the maximum number of the scaling based on the input data size, $\mathbf{A}_N^c$ is the approximation signal obtained at N level decomposition for each input channel, and $\mathbf{H}_s^c$, $\mathbf{V}_s^c$ and $\mathbf{D}_s^c$ (wavelet coefficients) are the horizontal, vertical and diagonal details respectively.

Utilizing the inverse Wavelet transform with details by neglecting the approximate data for each decomposition level, feature maps with increasing frequency bandwidth are created to represent salient regions from edge to textures, and the calculation of the feature maps is defined in (3) as follows [14]:

$$f_s^c\left(x,y\right) = \left(IWT_s\left(\left[\ \ \right], \mathbf{H}_s^c, \mathbf{V}_s^c, \mathbf{D}_s^c\right)\right)^2\qquad(3)$$

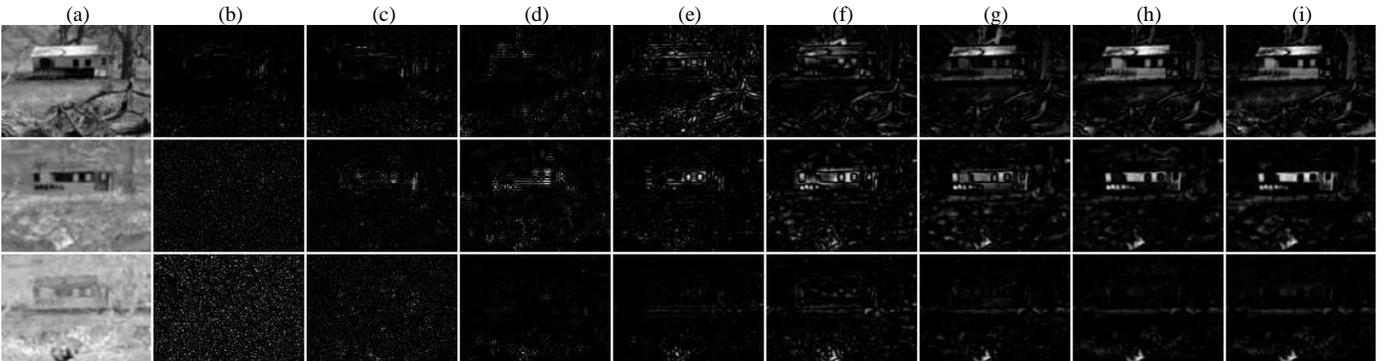

Fig. 3. (a) eigen vector space input channels (b)-(i) feature maps of input channels obtained by 1st level to 8th level WT details respectively as in (3)

where $f_s^c(x, y)$ is the feature map for the $s^{th}$ level decomposition obtained from the input data channel $c$, and $IWT_s(.)$ is the reconstruction function for the $s^{th}$ level features. In Fig. 3, some feature maps are shown for each eigenvector space input channel for the given input $RGB$ image of Fig.2 (b). These feature maps enable us to obtain saliency conspicuity map by summation of all features through each decomposition level for each input channel as in (4).

$$F^c(x, y) = N\left(\sum_s f_s^c(x, y)\right) \qquad (4)$$

where $F^c$ is the saliency conspicuity map of input channel $c$, and $N(.)$ is the function that consists of Gaussian smoothing and normalization to the range {0-1}. In Fig.4, corresponding saliency conspicuity maps are demonstrated.

It can be seen that each eigenvector space has their unique saliency representation where the $3^{rd}$ principal components have the most global saliency information when compared to the others. In addition, the $1^{st}$ eigenvector space provides more uniform salient regions which are advantageous if the object of interest has large size. Therefore, it is more appropriate to have the most importance on the $1^{st}$ transformed data's conspicuity map and the least affect from the $3^{rd}$ one during the fusion process when database includes images with large region of interest.

In addition to the local saliency model adopted from [14], we also use the saliency in [12] based on the contrast computed by the global mean value. The reason is to show that if the saliency data have lack of information or the saliency map has more false detections, the improvement on the saliency can be limited or even worse. The better the saliency model, the more improvement it can be achieved with the proposed integration model. In conventional case of [12], first, the $RGB$ color image is converted to CIE Lab color space which consists of luminance, $a$ and $b$ two color chromatic channels. Then, each conspicuity map is the result of difference between input channels pixel location and mean value of respective channel. Then, the absolute values of the feature vectors are taken as the saliency result for [12] on each image location. To make the analysis, the original algorithm is also adopted as in WT based case where the input channels are eigenvector space representations instead of any color space. Then, $m$-PCNN fusion integration to obtain saliency map can be completed on conspicuity maps instead of the absolute value of the feature vector as the final result. The comparisons and analysis are given in Section III.

### C. Fusion by m-PCNN

The next step is to combine all conspicuity maps to a final saliency map. The PCNN can be regarded as a promising algorithm on information fusion since its usability has been proven in many image processing applications [18, 19, 2, 17]. The common PCNN model is based on Eckhorn model which introduces the cat visual cortex [17]. The model is adjusted to PCNN for digital applications where the PCNN structure is given in Fig.5 [2, 18, 19].

And, the structure can be formulated by the following expressions [2, 17]:

$$F_{i,j}[n] = e^{-\alpha_F} F_{i,j}[n-1] + V_F\left(M * Y_{i,j}[n-1]\right) + I_{i,j} \qquad (5)$$

$$L_{i,j}[n] = e^{-\alpha_L} L_{i,j}[n-1] + V_L\left(W * Y_{i,j}[n-1]\right) \qquad (6)$$

$$U_{i,j}[n] = F_{i,j}[n]\left(1 + \beta L_{i,j}[n]\right) \qquad (7)$$

$$Y_{i,j}[n] = \begin{cases} 1, & U_{i,j}[n] > T_{i,j}[n-1] \\ 0, & else \end{cases} \qquad (8)$$

$$T_{i,j}[n] = e^{-\alpha_T} T_{i,j}[n-1] + V_T Y_{i,j}[n] \qquad (9)$$

where $F$ is the feeding compartment, $L$ is the linking compartment, $W$ and $M$ are the weights for the connections, $I$ is the input stimulus, the output by combining feeding and linking (7) depicts the internal state of the neuron at iteration $n$, $Y$ is the fired neurons that are defined by the dynamic threshold $T$, $V_F$ and $V_L$ are the used as scaling parameters, and $\alpha_F$ amd $\alpha_L$ are the time constants.

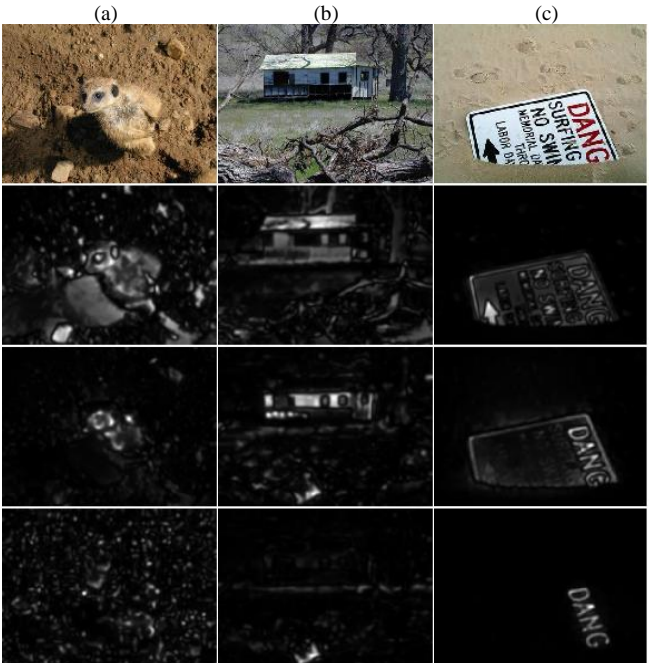

Fig. 4. Sample RGB images and their respective saliency conspicuity maps of each eigenvector space representations ($1^{st}$, $2^{nd}$, and $3^{rd}$ principal transformations respectively)

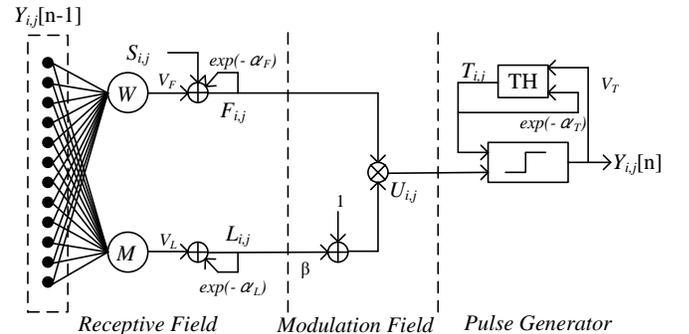

Fig. 5. Structure and work flow of a PCNN neuron

In this paper, due to the disadvantage of PCNN fusion for multi-channel cases and usefulness of $m$-PCNN [2] as stated earlier, $m$-PCNN with autonomous weighting adaptation is applied as the fusion model. The basic idea can be described by the fact that each input channel behaves as feeding compartment and also linkage of other input channels as in Fig.6 [2]. Also, the formulations for the structure are given as follows:

$$H_{i,j}^{k}[n] = e^{-\alpha_H} H_{i,j}^{k}[n-1] + V_H \left( W * Y_{i,j}[n-1] \right) + I_{i,j}^{k} \tag{10}$$

$$U_{i,j}[n] = \prod_{k=1}^{K} \left( 1 + \beta^{k} H_{i,j}^{k}[n] \right) \tag{11}$$

$$Y_{i,j}[n] = \begin{cases} 1, & U_{i,j}[n] > T_{i,j}[n-1] \\ 0, & else \end{cases} \tag{12}$$

$$T_{i,j}[n] = e^{-\alpha_T} T_{i,j}[n-1] + V_T Y_{i,j}[n] \tag{13}$$

where $k = \{1,2,3\}$ refers to the input channels (saliency conspicuity maps to be fused), $H$ is the external stimulus from the feed function in (10), $\beta_k$ is the weight of the $k^{th}$ data channel, and after the iterations are completed, the square root of $U$ will be taken as the fused final saliency to decrease the high variation during fusion due to the modulation. The initial values and parameters are given in the results section.

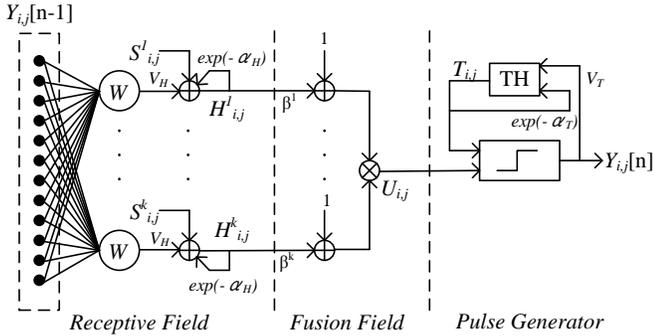

Fig. 6. Structure and work flow of a $m$-PCNN neuron

Using the eigenvalues based on the PCA, we can control the weight of each SCM autonomously during fusion where the 1st input channel has the highest weight since the SCM will be more representative by having more local and global information due to the high variation on the 1st eigenvector space. In addition, surrounding of the salient region will be more uniform, and noisy small false salient regions will be decreased or removed by the $m$-PCNN fusion model. The final saliency map examples are demonstrated in Fig.7 based on the SCMs in Fig.4.

## III. RESULTS & DISCUSSION

The purpose of the PCA and $m$-PCNN based fusion is to increase the uniformity around the salient region and to decrease the influence of irrelevant regions; therefore, the performance of the adopted saliency model is also important to have good results from the integration. If the saliency initially has good results, the proposed approach will certainly enhance the final saliency by making the local connection with appropriate parameter selections on appropriate regions. It should be noted that the aim is to demonstrate the affective fusion of conspicuity maps by PCA and $m$-PCNN utilization.

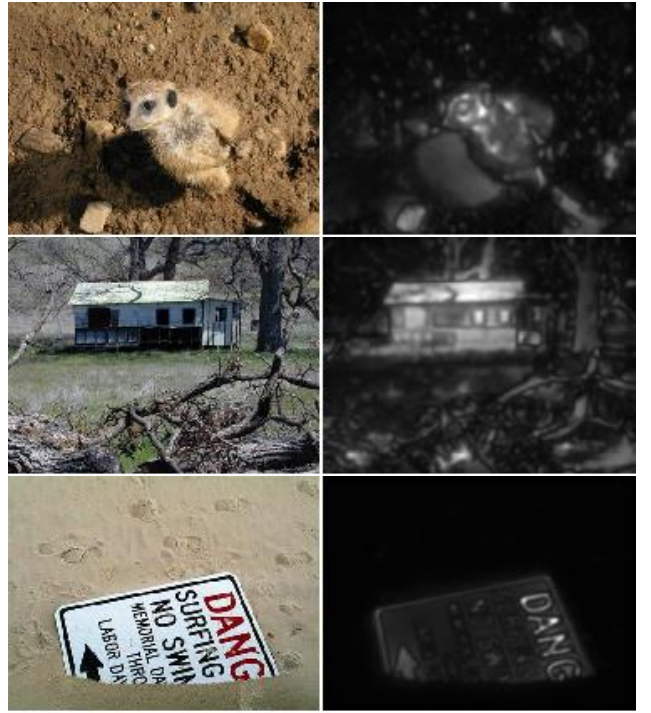

Fig. 7. Sample images and their saliency maps based on the proposed system

In this work, the parameters are chosen based on the experimental observations. Weighing matrix for the surrounding pixels $W$ in (10) is initialized by the pixel based distance (14) [19] with a size of 35x35 since the region of interests generally consists of large region.

$$W_{mn} = 1 / \left[ (m)^2 + (n)^2 \right] \tag{14}$$

where $m$ and $n$, representing the location of the surrounding pixels, are the distance to the centre pixel on x and y plane. Then, the other parameters are assigned as $\alpha_H = 0.001$ (10), $V_H = 15$ (10), $\alpha_T = 0.012$ (13) [2], $V_T = 100$ (13), $\beta_k = \varepsilon_k$, (11), $\varepsilon_k$ is the normalized eigenvalue of the $k^{th}$ eigenvector space in (2) where $\sum \varepsilon_k = 1$ should be satisfied. Initially, $Y_{i,j}$, $U_{i,j}$, and $T_{i,j}$ are all set to zero, and $H_{i,j}^{k}$ initially valued as $I_{i,j}^{k}$.

The evaluation of the final saliency map is accomplished by the consistency of salient region with the ground truth regions labeled by human subjects. By this way, the analysis on the saliency model can be observed quantitatively with the integration of PCA and $m$-PCNN to the WT based local saliency [14] and to the model in [12] as well. For this purpose, a subset of the Microsoft Image database (500 images) [21, 10] is used for experimental analysis. The dataset includes images and their respective ground truths from various subjects [21, 10].

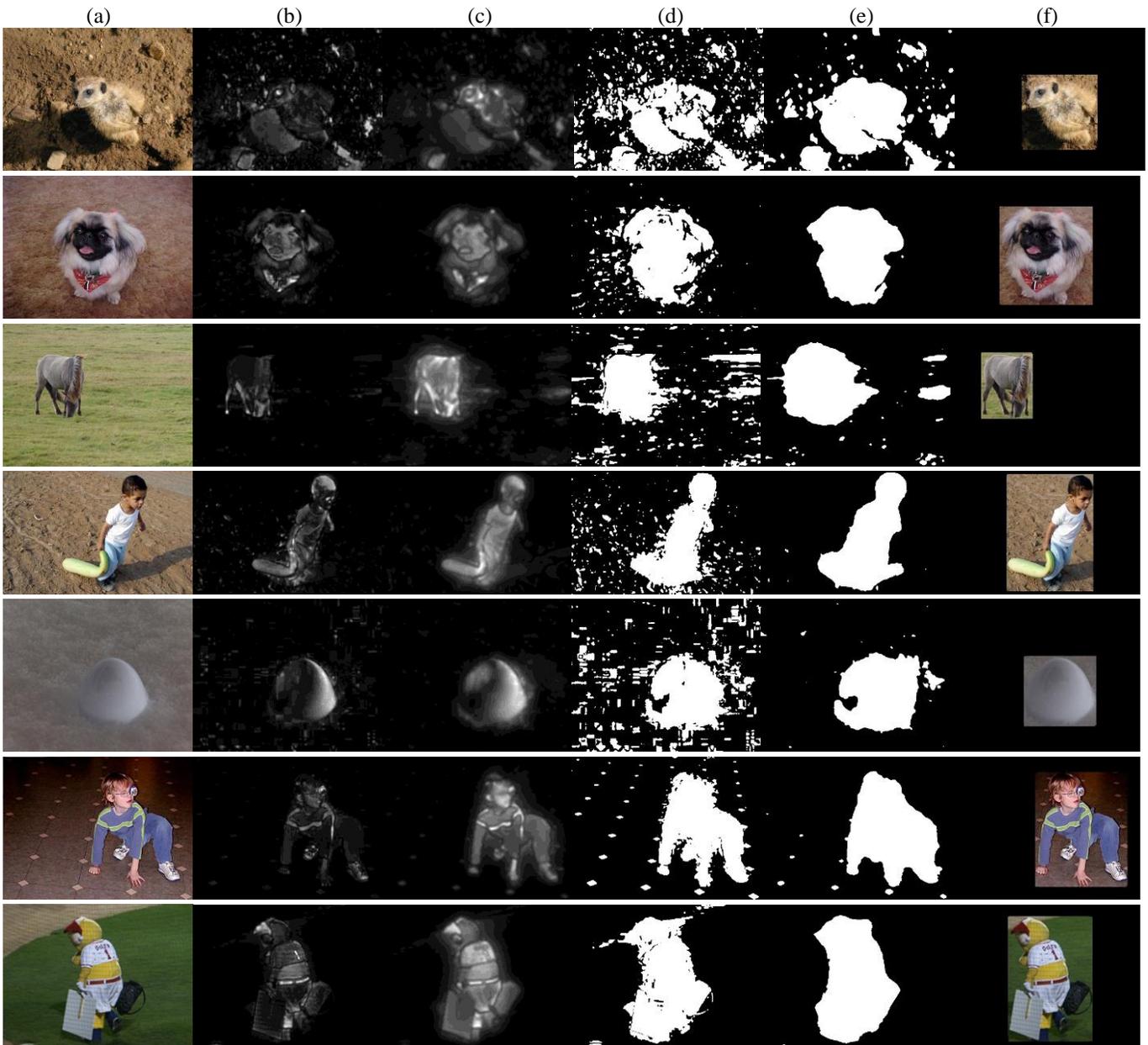

Fig. 8. (a) Sample images from the Microsoft Database, (b) WT based local saliency [14], (c) PCA and *m*-PCNN fusion based proposed saliency, (d) binary map generated from (b), (e) binary map generated from (c), (f) region of interest (ground truth) labeled by the subjects

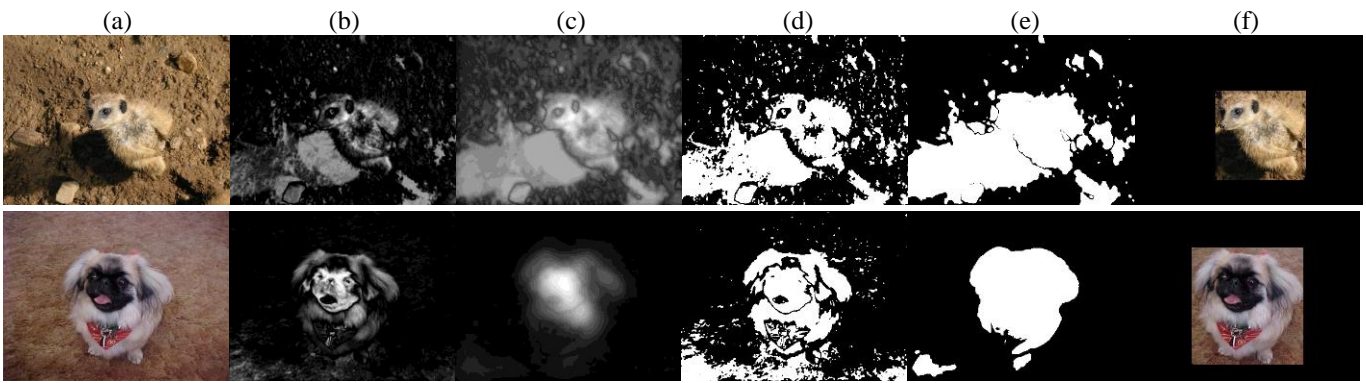

Fig. 9. (a) Sample images from the Microsoft Database, (b) saliency from [12], (c) PCA and *m*-PCNN fusion based integration for [12], (d) binary map generated from (b), (e) binary map generated from (c), (f) region of interest (ground truth) labeled by the subjects

The saliency maps obtained from the local saliency map in [14] and from the proposed model are also converted to binary maps for comparison to the ground truth in order to obtain a quantitative analysis. The binary map conversion is the result of a threshold operation with mean saliency value. In Fig.8, several examples are given for the local saliency model of [14] and the proposed structure with their respective binary maps. Same procedure is also employed for the analysis of [12] and its integration to PCA and $m$-PCNN. The outcomes from conventional saliency of [12] and integrated version of [12] are given in Fig.9.

In Fig.8(b), the local saliency model [14] makes fusion by taking the maximum value of the channel on each feature map, and applying the summation for all decomposition level contrary to (4) of the proposed structure. On the other hand, the difference of proposed saliency output in Fig.8(c) disables the channel fusion with maximum value but instead, PCA and $m$-PCNN based channel integration is applied to create the overall saliency map. As seen in Fig.8, the saliency values around the ground truth are enhanced with the proposed model, which can also be observed from the binary maps for each sample. The binary maps shown in Fig.8(e) have more uniformity and less irrelevant salient regions than the ones in Fig.8(d). This result is obtained because local and global contents are controlled by the PCA based input data transformation and because the surrounding pixels relations are taken into consideration during the channel fusion process due to the structure of $m$-PCNN.

On the other hand, regarding the integration of [12] with PCA and $m$-PCNN; however, if conspicuity maps are computed on eigenvector space input channels with more global based approach, $m$-PCNN will try to fuse the saliency information with less local information and more global content. Therefore, it is possible to have more misleading data from conspicuity maps to the $m$-PCNN. Obviously, in the 1st sample of Fig.9, there are improvements on the object of interest with the integration operation, but there are also increases in misdetections from [12] to its integration due to the higher influence of irrelevant salient regions. For the 2nd image in Fig.9, there are more consistencies on the saliency map from the conventional model of [12] compared to the ground truths region so integration of PCA and $m$-PCNN has better affect compared to the 1st sample image.

*Precision $P$, recall $R$, and F-Measure $F_\alpha$*, definitions are employed to enable quantitative analysis as defined below respectively [21,14]:

$$P = \frac{\sum_x \sum_y \big(g(x,y) \times s(x,y)\big)}{\sum_x \sum_y s(x,y)} \quad (15)$$

$$R = \frac{\sum_x \sum_y \big(g(x,y) \times s(x,y)\big)}{\sum_x \sum_y g(x,y)} \quad (16)$$

$$F_\alpha = \frac{(1+\alpha) \times P \times R}{\alpha \times P + R} \quad (17)$$

where the binary ground-truth map is symbolized by $g(x,y)$, the respective saliency binary map is given as $s(x,y)$, and $\alpha$ is a factor to determine the impact of *precision* over the *recall* while computing the *F-measure*. It is clear that *precision* is the ratio of correctly detected salient region to salient regions of the ground truth. *Recall* is the ratio of correctly detected salient regions to overall salient regions detected by the saliency model tested. And, *F-measure* is the harmonic mean of *precision* and *recall* as another performance criterion. In Table I, the improvement of the previous work [14] with current PCA and $m$-PCNN fusion integration is significant. Also, same improvements for these measurements are visible for [12] and its integration too. All the measurement criteria regarding the *precision*, *recall* and *F-measure* are increased. It demonstrates that the salient region representing the ground truth increased with some decrease in the irrelevant regions on final saliency map as shown in the examples in Fig.8 and the 2nd example of Fig.9.

*P*, *R*, and $F_\alpha$ measurements are quite useful when observing the change in the true positives (TP) cases in saliency enhancement since both precision and recall values are directly related to the detected salient regions. Even though, *R* can give some analysis on the false positives (FP) of the detected salient regions, these measurements are not good enough to express the changes in both TP and FP. Therefore, we also utilize the measurement called area under curve (AUC) which is also given in Table I [22, 23, 19]. AUC is calculated as the area under the receiver operating characteristics (ROC) curve that is defined by true positive rate (TPR) with respect to false positive rate (FPR) by changing the threshold value [19, 23]. Thus, it is another validation for prediction analysis. The function used to compute AUC is from the work [24] where the detailed information can be found. AUC results for the proposed model also expresses the improvement on the previous work [14] with the proposed fusion based on PCA and $m$-PCNN model.

TABLE I
QUANTITATIVE MEASUREMENTS

| Method | $P$ | $R$ | $F_\alpha$ | AUC |
|---|---|---|---|---|
| *WT based local saliency in [14]* | 0.6180 | 0.5911 | 0.5967 | 0.7253 |
| *Local saliency in [14] with proposed fusion model* | 0.6334 | 0.6796 | 0.6283 | 0.7565 |
| *Conventional saliency method in [25]* | 0.5312 | 0.4612 | 0.4957 | 0.6473 |
| *saliency in [25] with proposed fusion model* | 0.6156 | 0.6128 | 0.5957 | 0.7145 |

*P*, *R*, and $F_\alpha$ values improved for the conspicuity maps for [12] with proposed structure since it will be expected if the region size for true positive region increases. However, as stated before, these measurements cannot express the FPR cases directly in which Table I shows the decrease in AUC due to the misleading based on the conspicuity maps that can be onserved from 1st example of Fig.9. These results of [12] and its integrations are the reasons to select a saliency model with local and global content for adaptation to proposed fusion model. This is why, local saliency in previous work [14] is selected as the proposed model for this work; in addition, it also includes local to global saliency information on conspicuity maps from the multi-resolution WT based computation. And, work [12] and its integration are chosen to make analysis and comparison to the proposed system with [14]. The overall analysis demonstrates the effectiveness of the fusion model with the selected WT based local conspicuity map model.

## IV. Conclusion

In this paper, eigenvector space saliency conspicuity maps are fused with *m*-PCNN model with local to global local saliency model based on Wavelet transform. The proposed system aims to improve the uniformity around the salient regions and to decrease irrelevant small salient regions based on PCA and *m*-PCNN fusion process. From the RGB image, three transformed image channels are created by PCA using the eigenvectors of the principal components. Then, conspicuity maps are computed from these channel with the defined WT based approach to utilize the m-PCNN fusion model with the weighting control of input channels autonomously by using the eigenvalues of each eigenvector space. Thus, local to global information can be controlled up to some extent during the fusion of conspicuity maps to obtain saliency map. Therefore, both fusion of conspicuity maps and improved saliency computation is achieved using PCA and m-PCNN integration to a state of the art model. Experimental results and quantitative analysis have promising results for future works related to the method by improving the saliency map during the fusion process.